\pdfoutput=1

\documentclass[11pt]{article}

\usepackage{ACL2023}

\usepackage{times}
\usepackage{latexsym}

\usepackage[T1]{fontenc}

\usepackage[utf8]{inputenc}

\usepackage{microtype}

\usepackage{inconsolata}
\usepackage{graphicx}
\usepackage{algorithm}
\usepackage[noend]{algorithmic}
\usepackage{amsfonts}
\usepackage{amsmath}
\usepackage{bm}
\usepackage{arydshln}
\usepackage{soul}
\usepackage{bbding}

\usepackage{amsmath,amssymb,amsfonts}

\setul{0.2ex}{0.1ex}

\title{\emph{Information Screening whilst Exploiting!} Multimodal Relation Extraction with Feature Denoising and Multimodal Topic Modeling\Thanks{ The work is substantially supported by Alibaba Group through the Alibaba Innovative Research (AIR) Program, and also partially supported by the Sea-NExT Joint Lab at the National University of Singapore.}}

\author{
Shengqiong Wu\textsuperscript{\rm 1},  \, 
Hao Fei\textsuperscript{\rm 1}\Thanks{ Corresponding author: Hao Fei}, \, 
Yixin Cao\textsuperscript{\rm 2},  \, 
Lidong Bing\textsuperscript{\rm 3},  \, 
Tat-Seng Chua\textsuperscript{\rm 1} \\
\textsuperscript{\rm 1} Sea-NExT Joint Lab, School of Computing, National University of Singapore \\
\textsuperscript{\rm 2} Singapore Management University, \quad
\textsuperscript{\rm 3} DAMO Academy, Alibaba Group  \\
\tt {swu@u.nus.edu \quad haofei37@nus.edu.sg   \quad  caoyixin2011@gmail.com} \\ 
\tt {l.bing@alibaba-inc.com   \quad  dcscts@nus.edu.sg}
}

\begin{document}
\maketitle

\begin{abstract}
Existing research on multimodal relation extraction (MRE) faces two co-existing challenges, \emph{internal-information over-utilization} and \emph{external-information under-exploitation}.
To combat that, we propose a novel framework that simultaneously implements the idea of \emph{internal-information screening} and \emph{external-information exploiting}.
First, we represent the fine-grained semantic structures of the input image and text with the visual and textual scene graphs, which are further fused into a unified cross-modal graph (CMG).
Based on CMG, we perform structure refinement with the guidance of the graph information bottleneck principle, actively denoising the less-informative features.
Next, we perform topic modeling over the input image and text, incorporating latent multimodal topic features to enrich the contexts.
On the benchmark MRE dataset, our system outperforms the current best model significantly.
With further in-depth analyses, we reveal the great potential of our method for the MRE task.
Our codes are open at \url{https://github.com/ChocoWu/MRE-ISE}.

\end{abstract}

\section{Introduction}

Relation extraction (RE), determining the semantic relation between a pair of subject and object entities in a given text \cite{YuXZLWW20}, has played a vital role in many downstream natural language processing (NLP) applications, e.g., knowledge graph construction \cite{WangLLBL19,mondal-etal-2021-end}, question answering \cite{cao-etal-2022-program}.
But in realistic scenarios (i.e., social media), data is often in various forms and modalities (i.e., texts, images), rather than pure texts.
Thus, multimodal relation extraction has been introduced recently \cite{ZhengWFF021}, where additional visual sources are added to the textual RE as an enhancement to the relation inference. 
The essence of a successful MRE lies in the effective utilization of multimodal information.
Certain efforts have been made in existing MRE work and achieved promising performances, where delicate interaction and fusion mechanisms are designed for encoding the multimodal features \cite{ZhengFFCL021,ChenZLYDTHSC22,ChenZLDTXHSC22}.
Nevertheless, current methods still fail to sufficiently harness the feature sources from two information perspectives, which may hinder further task development.

\begin{figure}[!t]
\centering
\includegraphics[width=0.98\columnwidth]{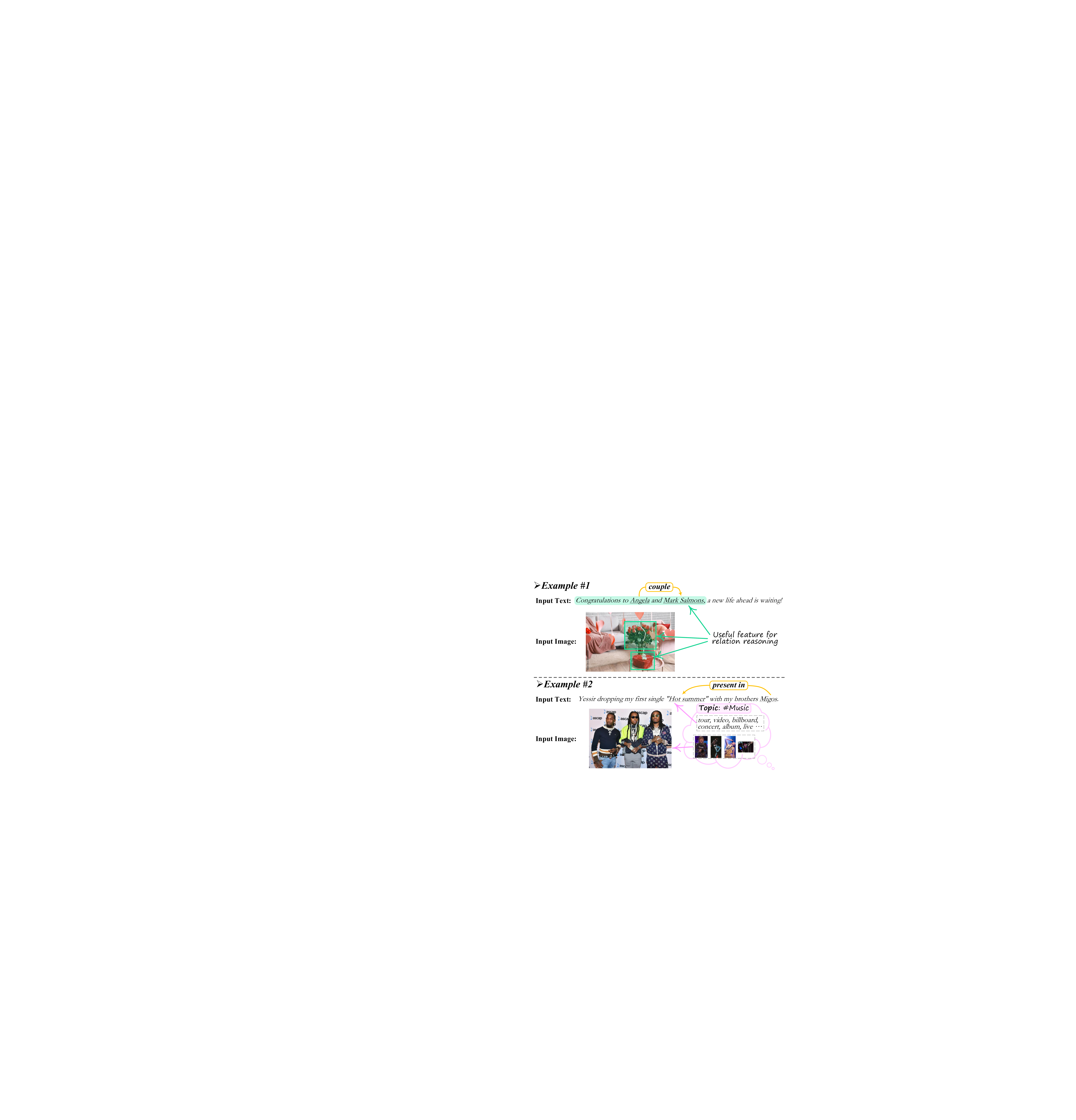}
\vspace{-1mm}
\caption{
Examples of multimodal relation extraction (MRE). 
In texts, the subject and object entities are underlined, and linked by the relational tags.
}
\label{fig:intro}
\vspace{-3mm}
\end{figure}

\textbf{\emph{Internal-information over-utilization.}}
On the one hand, most existing MRE methods progressively incorporate full-scale textual and visual sources into the learning, under the assumption that all the input information certainly contributes to the task. 
In fact, prior textual RE research extensively shows that only parts of the texts are useful to the relation inference \cite{YuXZLWW20}, and accordingly propose to prune over input sentences \cite{Zhang0M18}.
The case is more severe for the visual inputs, as not all and always the visual sources play positive roles, especially on the social media data.
As revealed by \citet{vempala-preotiuc-pietro-2019-categorizing}, as high as 33.8\% of visual information serves no context or even noise in MRE.
\citet{0023HDWSSX22} thus propose to selectively remove images from the input image-text pairs.
Unfortunately, such coarse-grained instance-level filtering largely hurts the utility of visual features.
We argue that a fine-grained feature screening over both the internal image and text features is needed.
Taking the example \#1 in Fig. \ref{fig:intro}, the textual expressions `\emph{Congratulations to Angela and Mark Salmons}' and the visual objects of `\emph{gift}' and `\emph{roses}' are valid clues to infer the `\emph{couple}' relation between `
\emph{Angela}' and `\emph{Mark Salmons}', while the rest of text and visual information is essentially the task-irrelevant noise.

\textbf{\emph{External-information under-exploitation.}}
On the other hand, although compensating the text inputs with visual sources, there can be still information deficiency in MRE, in particular when the visual features serve less (or even negative) utility.
This is especially the case for social media data, where the contents are less-informative due to the short text lengths and low-relevant images \cite{baly-etal-2020-written}.
For the example \#2 in Fig. \ref{fig:intro},
due to the lack of necessary contextual information, it is tricky to infer the relation `\emph{present in}' between `\emph{Hot summer}' (an album name) and `\emph{Migos}' (a singer name) based on both the image and text sources.
In this regard, more external information should be considered and exploited for MRE.
Fortunately, the topic modeling technique offers a promising solution, which has been shown to enrich the semantics of the raw data, and thus facilitate NLP applications broadly \cite{zeng-etal-2018-topic}.
For the above same example, if an additional `\emph{music}' topic feature is leveraged into the context, the relation inference can be greatly eased.

Taking into account the above two observations, in this work, we propose a novel framework to improve MRE.
As shown in Fig. \ref{fig:framework}, we first employ the scene graphs (SGs) \cite{JohnsonKSLSBL15} to represent the input vision and text, where SGs advance in intrinsically depicting the fine-grained semantic structures of texts or images.
We fuse both the visual and textual SGs into a cross-modal graph (CMG) as our backbone structure.
Next, we reach the first goal of internal-information screening by adjusting the CMG structure via the graph information bottleneck (GIB) principle \cite{WuRLL20}, i.e., \emph{GIB-guided feature refinement}, during which the less-informative features are filtered and the task-relevant structures will be highlighted.
Then, to realize the second goal of external-information exploiting, we perform \emph{multimodal topic integration}.
We devise a latent multimodal topic module to produce both the textual and visual topic features based on the multimodal inputs.
The multimodal topic keywords are integrated into the CMG to enrich the overall contexts, based on which we conduct the final reasoning of relation for input.

\begin{figure}[!t]
\centering
\includegraphics[width=0.98\columnwidth]{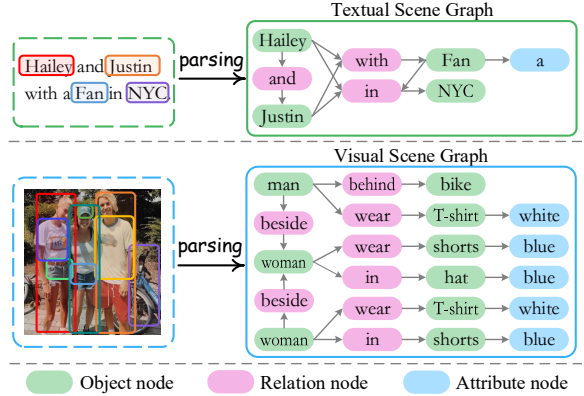}
\vspace{-1mm}
\caption{
Examples of textual and visual scene graphs.
}
\label{fig:scene}
\vspace{-2mm}
\end{figure}

We perform experiments on the benchmark MRE dataset \cite{ZhengFFCL021}, where the results show that our framework significantly boosts the current state of the art.
Further analyses demonstrate that the GIB-guided feature refinement helps in effective input denoising, 
and the latent multimodal topic module induces rich task-meaningful visual\&textual topic features as extended contexts.
We finally reveal that the idea of internal-information screening is especially important to the scenario of higher text-vision relevance, while the external-information exploiting particularly works for the lower text-vision relevance case.

To sum up, this work contributes by introducing a novel idea of simultaneous information subtraction and addition for multimodal relation extraction.
The internal-information over-utilization and external-information under-exploitation are two common co-existing issues in many multimodal applications, to which our method can be broadly applied without much effort.

\section{Preliminary}

\vspace{-1mm}
\subsection{Textual and Visual Scene Graph}

There have been the visual scene graph (VSG) \cite{JohnsonKSLSBL15} and textual scene graph (TSG) \cite{wang-etal-2018-scene},
where both of them include three types of nodes: \emph{object node}, \emph{attribute node}, and \emph{relationship node}.
All the nodes come with a specific label text, as illustrated in Fig. \ref{fig:scene}.
In an SG, object and attribute nodes are connected with other objects via pairwise relations.
As intrinsically describing the semantic structures of scene contexts for the given texts or images, SGs are widely utilized as types of external features integrated into downstream applications for enhancements, e.g., image retrieval \cite{JohnsonKSLSBL15}, image generation \cite{JohnsonGF18} and image captioning \cite{YangTZC19}.
We also take advantage of these SG structures for better cross-modal semantic feature learning.
Formally, we define a scene graph as ${G}$=$({V},{E})$, where ${V}$ is the set of nodes, and ${E}$ is the set of edges.

\vspace{-1mm}
\subsection{Graph Information Bottleneck Principle}
\label{GIBP}

The information bottleneck (IB) principle \cite{AlemiFD017} is designed for information compression.
Technically, IB learns a minimal feature $Z$ to represent the raw input $X$ that is sufficient to infer the task label $Y$.
Further, the graph-based IB has been introduced for the graph data modeling \cite{WuRLL20}, i.e., by refining a raw graph ${G}$ into an informative yet compact one ${G}^{-}$, by optimizing:
\setlength\abovedisplayskip{2pt}
\setlength\belowdisplayskip{2pt}
\begin{equation}\small\label{GIB}
    \mathop{min}\limits_{{G}^{-}} \, [ - I({G}^{-}, Y) + \beta \cdot I({G}^{-}, {G}) ] \,,
\end{equation}
where $I({G}^{-}, {G})$ minimizes the mutual information between ${G}$ and ${G}^{-}$ such that ${G}^{-}$ learns to be the minimal and compact one of ${G}$.
$I({G}^{-}, Y)$ is the prediction objective, which encourages ${G}^{-}$ to be informative enough to predict the label $Y$.
$\beta$ is a Lagrangian coefficient.
We will employ the GIB principle for internal-information screening.

\vspace{-1mm}
\subsection{Latent Multimodal Topic Modeling}
\label{sec:Latent Multimodal Topic Modeling}

We introduce a \ul{la}tent \ul{m}ultimodal t\ul{o}pic (\textsc{Lamo}) model.
Technically, we first represent the input text $T$ with a bag-of-word (BoW) feature $\bm{b}^{T}$, and represent image $I$ with a visual BoW (VBoW)\footnote{
Note that visual topic words are visual objects.
} $\bm{b}^{I}$.
The topic generative process is described as follows:\\
\indent $\bullet$ Draw a topic distribution $\bm{\theta} \sim \mathcal{N}(\bm{\mu}, \bm{\sigma})$.\\
\indent $\bullet$ For each word token $w^{T}_i$ and visual token $w^{I}_j$:\\
\indent \indent $\circ$ Draw $w^{T}_i \sim Multinomial(\bm{\chi}, \bm{\theta})$,\\
\indent \indent $\circ$ Draw $w^{I}_j \sim Multinomial(\bm{\psi}, \bm{\theta})$.\\
where $\bm{\mu}$ and $\bm{\sigma}$ are the mean and variance vector for the posterior probability $p(\bm{\theta}|T, I)$.
$\bm{\chi} \in \mathbb{R}^{K \times U^{T}}$ and $\bm{\psi} \in \mathbb{R}^{K \times U^{I}}$ are the probability matrices of \emph{textual topic-word} and \emph{visual topic-word}, respectively.
$K$ is the pre-defined topic numbers, and $U^{T}$ and $U^{I}$ are textual and visual vocabulary size.

\begin{figure}[!t]
\centering
\includegraphics[width=0.98\columnwidth]{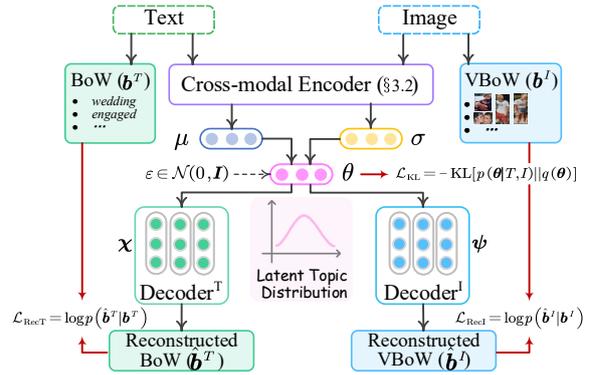}
\vspace{-1mm}
\caption{The schematic of our latent multimodal topic (\textsc{Lamo}) model.
}
\label{fig:mtd}
\vspace{-2mm}
\end{figure}

As depicted in Fig. \ref{fig:mtd}, $\bm{\mu}$ and $\bm{\sigma}$ are produced from a cross-modal feature encoder upon $T$ and $I$.
The topic distribution is yielded via $\bm{\theta}$=$\text{Softmax}(\bm{\mu}$+$\bm{\sigma} \cdot \varepsilon)$, where $\varepsilon \in \mathcal{N}(0, \bm{I})$.
Then, we autoregressively reconstruct the input $\bm{b}^T$ and $\bm{b}^I$ based on $\bm{\theta}$:
\setlength\abovedisplayskip{2pt}
\setlength\belowdisplayskip{2pt}
\begin{equation}\small\label{chi}
p(\bm{b}_i^T|\bm{\chi}, \bm{\theta}) = \text{Softmax}(\bm{\theta} \cdot \bm{\chi} | \bm{b}_{<i}^T )\,,
\end{equation}
\begin{equation}\small\label{psi}
p(\bm{b}_i^I|\bm{\psi}, \bm{\theta}) = \text{Softmax}(\bm{\theta} \cdot \bm{\psi} | \bm{b}_{<i}^I)\,.
\end{equation}
Then, with the activated $k$-th topic (via argmax over $\bm{\theta}$), we obtain the distributions of the textual and visual topic words by slicing the $\bm{\chi}[k,:] \in \mathbb{R}^{U^{T}}$ and $\bm{\psi}[k,:] \in \mathbb{R}^{U^{I}}$.

As shown in Fig. \ref{fig:mtd}, the objective of topic modeling is derived as follows:
\begin{equation}\label{eq-Lamo}
\small
\begin{aligned}
\setlength\abovedisplayskip{2pt}
\setlength\belowdisplayskip{2pt}
\mathcal{L}_{\text{\scriptsize LAMO}} = \mathcal{L}_{\text{\scriptsize KL}} + \mathcal{L}_{\text{\scriptsize RecT}} +\mathcal{L}_{\text{\scriptsize RecI}} \,.
\end{aligned}
\end{equation}
Appendix $\S$\ref{Latent Multimodal Topic Modeling} extends the description of \textsc{Lamo}.

\begin{figure*}[!ht]
\centering
\includegraphics[width=0.99\textwidth]{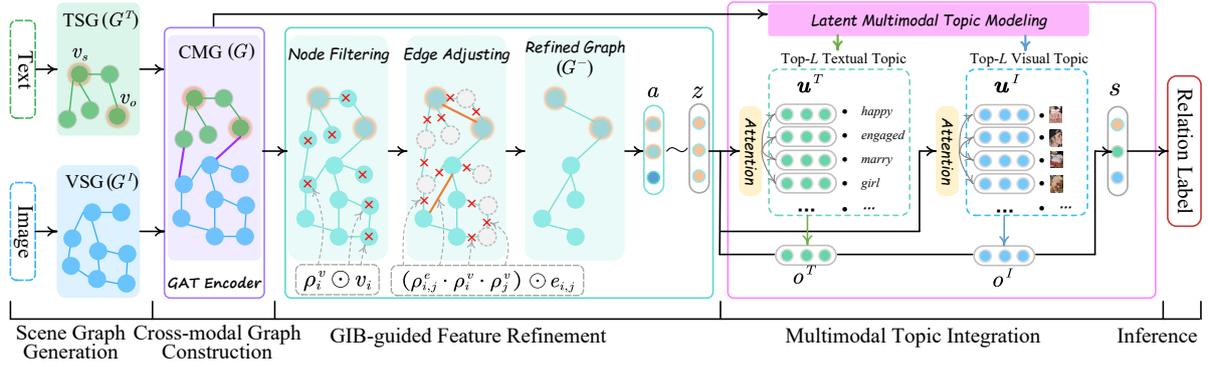}
\vspace{-1mm}
\caption{
Overview of our proposed framework. 
}
\label{fig:framework}
\vspace{-2mm}
\end{figure*}

\section{MRE Framework}
\vspace{-1mm}

As shown in Fig. \ref{fig:framework}, our overall framework consists of five tiers.
First, the model takes as input an image $I$ and text $T$, as well as the subject ${v}_{s}$ and object entity ${v}_{o}$.
We represent $I$ and $T$ with the corresponding VSG and TSG.
Then, the VSG and TSG are assembled as a cross-modal graph, which is further modeled via a graph encoder.
Next, we perform GIB-guided feature refinement over the CMG for internal-information screening, which results in a structurally compact backbone graph.
Afterwards, the multimodal topic features induced from the latent multimodal topic model are integrated into the previously obtained feature representation for external-information exploitation.
Finally, the decoder predicts the relation label $Y$ based on the enriched features.

\vspace{-1mm}
\subsection{Scene Graph Generation}
\label{Scene Graph Generation}

We employ the off-the-shelf parsers to generate the VSG (i.e., ${G}^{I}$=$({V}^{I}, {E}^{I})$) and TSG (i.e.,
$G^{T}$=$(V^{T}, {E}^{T})$), respectively.
We denote the representations of VSG nodes as $\bm{X}^{I}$=$\{\bm{x}_1^{I}, \cdots, \bm{x}_n^{I}\}$, where each node embedding $\bm{x}_i^{I}$ is the concatenation of the object region representations and the corresponding node label embeddings.
We directly represent the TSG nodes as $\bm{X}^{T}$=$\{\bm{x}_1^{T}, \cdots, \bm{x}_m^{T}\}$, where each $\bm{x}_j^{T}$ is the contextualized word embedding.
Note that both visual objects and text token representations are obtained from the CLIP \cite{RadfordKHRGASAM21} encoder, which ensures an identical embedding space across the two modalities. 
More details are provided in the appendix \S\ref{scene graph generating}\&\S\ref{visual node embedding}.

\vspace{-1mm}
\subsection{Cross-modal Graph Construction}
\label{Cross-modal Graph Construction}

Next, we consider merging the VSG and LSG into one unified backbone cross-modal graph (CMG).
Let's denote CMG as ${G}$=$({V},{E})$, where ${V}$(=${V}^{T} \cup {V}^{I}$) is the union of ${V}^I$ and ${V}^T$.
${E}$(=${E}^{T} \cup {E}^{I} \cup {E}^{\times}$) is the set of edges, including the \emph{intra-modal} edges (${E}^{I}$ and ${E}^{T}$), and \emph{inter-modal} hyper-edges ${E}^{\times}$.
To build the cross-modal hyper-edges between each pair of VSG node $v^I_i$ and TSG node $v^T_j$, we measure the relevance score $s$ in between:
\begin{equation}\small
\setlength\abovedisplayskip{2pt}
\setlength\belowdisplayskip{2pt}
    s_{v^I_i,v^T_j}=cos(\bm{x}_i^{I}, \bm{x}_j^{T}) \,.
\end{equation}
A hyper-edges $e^{\times}_{i,j}$ is created if $s_{v^I_i,v^T_j}$ is larger than a pre-defined threshold $\lambda$.
Node representations from VSG and TSG are copied as the CMG's node representations, i.e., $\bm{X}$=$\bm{X}^T\cup \bm{X}^I$.
We denote each edge $e_{i,j}(\in E)$=1 if there is an edge between nodes, and $e_{i,j}$=0 and vice versa.
Next, a graph attention model \citep[GAT; ][]{VelickovicCCRLB18} is used to fully propagate the CMG:
\begin{equation}\small
\setlength\abovedisplayskip{2pt}
\setlength\belowdisplayskip{2pt}
    \bm{H} = \{\bm{h}_1, \cdots, \bm{h}_{m+n}\} = \text{GAT}(G, \bm{X}) \,.
\end{equation}

\vspace{-1mm}
\subsection{GIB-guided Feature Refinement}
\label{GIB-guided Feature Refinement}

In this step, we propose a \ul{G}IB-guided f\ul{e}ature refi\ul{ne}ment (\textsc{Gene}) module to optimize the initial CMG structure such that we fine-grainedly prune the input image and text features.
Specifically, with the GIB guidance, we 1) filter out those task-irrelevant nodes, and 
2) adjust the edges based on their relatedness to the task inference.

\vspace{-1mm}
\paragraph{Node Filtering}
We assign a 0 or 1 value $\rho_i^{v}$ to a node $v_i$ indicating whether to prune or keep $v_i$, i.e., via $\rho_i^{v} \odot v_i$.
We sample the value from the \emph{Bernoulli} distribution, i.e.,
$\rho_i^{v}$$\in$$\{0,1\}$$\sim$$\text{Bernoulli}(\pi_i^{v})$, where $\pi_i^{v}$$\in$$(0,1)$ is a parameter.
While the sampling is a discrete process, we make it differentiable via the concrete relaxation method \cite{JangGP17}:
\begin{equation}\small\label{rho-v}
\setlength\abovedisplayskip{2pt}
\setlength\belowdisplayskip{2pt}
    \rho_i^{v} = \text{Sigmoid}(\frac{1}{\tau}(\text{log}\frac{\pi_i^{v}}{1-\pi_i^{v}} + \text{log}\frac{\epsilon}{1-\epsilon})) \,,
\end{equation}
where $\tau$ is the temperature, 
$\epsilon \sim \text{Uniform}(0, 1)$. 
We estimate $\pi_i^{v}$ by considering both the $v_i$'s \emph{$l$-order context} and the influence of target entity pair:
\begin{equation} \small
\setlength\abovedisplayskip{3pt}
\setlength\belowdisplayskip{3pt}
\begin{aligned}\label{pi-v}
&\bm{r}^v_i = \text{Att} ( v_i, \varphi(v_i)  ;\bm{H} ) \,, \\
\pi_i^{v} &= \text{Sigmoid}(\text{FFN}([ \bm{r}^v_i ; \bm{h}_{s} ; \bm{h}_{o} ])) \,, 
\end{aligned}
\end{equation}
where Att($\cdot$) is an attention operation,
$\varphi(v_i)$ is the $l$-order neighbor nodes of $v_i$, $\bm{h}_{s}$ and $\bm{h}_{o}$ are the representations of the subject and object entity.

\vspace{-1mm}
\paragraph{Edge Adjusting}
Similarly, we take the same sampling operation (Eq. \ref{rho-v}) to generate a signal $\rho_{i,j}^{e}$ for any edge $e_{i,j}$, during which we also consider the \emph{$l$-order context} features and the target entity pair:
\setlength\abovedisplayskip{4pt}
\setlength\belowdisplayskip{4pt}
\begin{equation} \small
\label{pi-e}
\begin{aligned}
\bm{r}^e_{i,j} &= \text{Att} ( v_i, \varphi(v_i), v_j, \varphi(v_j)  ;\bm{H} ) \,, \\
\pi_{i,j}^{e} &= \text{Sigmoid}(\text{FFN}([ \bm{r}^e_{i,j} ; \bm{h}_{s} ; \bm{h}_{o} ])) \,, 
\end{aligned}
\end{equation}
where $\varphi(v_i)$ and $\varphi(v_j)$ are the $l$-order neighbor nodes of $v_i$ and $v_j$.
Instead of directly determining the existence of $e_{i,j}$ with $\rho_{i,j}^{e}$, we also need to take into account the existences of $v_i$ and $v_j$, i.e., $(\rho_{i,j}^{e} \cdot \rho_i^{v} \cdot  \rho_j^{v}) \odot e_{i,j}$.
Because even if $\rho_{i,j}^{e}$=1, an edge is non-existent when its affiliated nodes are deleted.

Thereafter, we obtain an adjusted CMG, i.e., $G^{-}$, which is further updated via the GAT encoder, resulting in new node representations $\bm{H}^{-}$.
We apply pooling operation on $\bm{H}^{-}$ to obtain the overall graph presentation $\bm{g}$, which is concatenated with two entity representations as the context feature $\bm{a}$:
\begin{equation}\label{final-context-feature-a}
\setlength\abovedisplayskip{2pt}
\setlength\belowdisplayskip{2pt}
  \bm{a} = [\bm{g}; \bm{h}_{s} ; \bm{h}_{o}] \,.
\end{equation}

\vspace{-1mm}
\paragraph{GIB Optimization}

To ensure that the above-adjusted graph $G^{-}$ is sufficiently informative (i.e., not wrongly pruned), we consider a GIB-guided optimization.
We denote $\bm{z}$ as the compact information of the resulting $G^{-}$, which is sampled from a Gaussian distribution parameterized by $\bm{a}$.
Then, we rephrase the raw GIB objective (Eq. \ref{GIB}) as:
\begin{equation}\small\label{GIB-total}
\setlength\abovedisplayskip{2pt}
\setlength\belowdisplayskip{2pt}
\mathcal{L}_{\text{\scriptsize GIB}} = \mathop{min}\limits_{\bm{z}} \, [ - I(\bm{z}, Y) + \beta \cdot I(\bm{z}, {G}) ] \,.
\end{equation}
The first term $-I(\bm{z}, Y)$ can be expanded as:
\begin{small}
\begin{equation}\label{GIB5}
\begin{aligned}\small
\setlength\abovedisplayskip{2pt}
\setlength\belowdisplayskip{2pt}
    -I(\bm{z}, Y) &\leq -\! \int\! p(Y, \bm{z})\log q(Y|\bm{z}) dY d\bm{z} + H(Y) \\
    & := \mathcal{L}_{\text{\scriptsize CE}}(q(Y|\bm{z}), Y) \,,
\end{aligned}
\end{equation}
\end{small}
where $q(Y|\bm{z})$ is a variational approximation of the true posterior $p(Y, \bm{z})$.
For the second term $I(\bm{z}, {G})$, we estimate its upper bound via reparameterization trick \cite{KingmaW13}:
\begin{equation}\label{eq-GIB3}
\small
\setlength\abovedisplayskip{2pt}
\setlength\belowdisplayskip{2pt}
\begin{aligned}
I(\bm{z}, {G}) &\leq \int p(\bm{z}| {G})\log\frac{p(\bm{z}|{G})}{r(\bm{z})} d\bm{z} d{G}  \\
& := \text{KL}(p(\bm{z}|{G})||r(\bm{z})) \,.
\end{aligned}
\end{equation}
We run \textsc{Gene} several iterations for sufficient refinement.
In Appendix $\S$\ref{GIB-process} we detail all the technical processes of GIB-guided feature refinement.

\vspace{-2mm}
\subsection{Multimodal Topic Integration}

\vspace{-1mm}
We further enrich the compressed CMG features with more semantic contexts, i.e., the multimodal topic features.
As depicted in Sec. $\S$\ref{sec:Latent Multimodal Topic Modeling}, our \textsc{Lamo} module takes as input the backbone CMG representation $\bm{H}$ and induces both the visual and textual topic keywords that are semantically relevant to the input content.
Note that we only retrieve the associated top-$L$ textual and visual keywords, separately.
Technically, we devise an attention operation to integrate the embeddings of the multimodal topic words ($\bm{u}^{T}$ and $\bm{u}^{I}$, from CLIP encoder) into the resulting feature representation $\bm{z}$ of \textsc{Gene}:
\setlength\abovedisplayskip{2pt}
\setlength\belowdisplayskip{2pt}
\begin{equation}\small\label{att-weights}
\begin{aligned}
\alpha^{T/I}_i &= \frac{\exp( \text{FFN} ([ \bm{u}^{T/I}_i ; \bm{z} ] ))}{\sum_i^{L} \exp( \text{FFN} ([ \bm{u}^{T/I}_i ; \bm{z} ] )) } \,, \\
\bm{o}^{T/I} &= \sum_i^{L} \alpha^{T/I}_i  \bm{u}^{T/I}_i \,.
\end{aligned}
\end{equation}
We finally summarize these three representations as the final feature:
\begin{equation}\small
\setlength\abovedisplayskip{2pt}
\setlength\belowdisplayskip{2pt}
\bm{s} = [\bm{z} ; \bm{o}^{T} ; \bm{o}^{I}]\,.
\end{equation}

\vspace{-2mm}
\subsection{Inference and Learning}
\label{Inference and Learning}

\vspace{-1mm}
Based on $\bm{s}$, a softmax function predicts the relation label $\hat{Y}$ for the entity pair $v_s$\&$v_o$.
The training of our overall framework is based on a warm-start strategy.
First, \textsc{Gene} is trained via $\mathcal{L}_{\text{\scriptsize GIB}}$ (Eq. \ref{GIB-total}) for learning the sufficient multimodal fused representation in CMG, and refined features from compacted CMG.
Then \textsc{Lamo} module is unsupervisedly pre-trained separately via $\mathcal{L}_{\text{\scriptsize LAMO}}$ (Eq. \ref{eq-Lamo}) on the well-learned multimodal fused representations so as to efficiently capture the task-related topic.
Once the two modules have converged, we train our overall framework with the final cross-entropy task loss $\mathcal{L}_{\text{\scriptsize CE}}(\hat{Y}, Y)$, together with the above two learning loss:
\begin{equation}\small\label{loss-all}
    \mathcal{L} = \mathcal{L}_{\text{\scriptsize CE}} + \eta_1 \mathcal{L}_{\text{\scriptsize GIB}} + \eta_2 \mathcal{L}_{\text{\scriptsize LAMO}} \,.
\end{equation}

\begin{table}[!t]
\fontsize{9}{10.5}\selectfont
 \setlength{\tabcolsep}{3mm}
\begin{center}
\begin{tabular}{lcccc} 
\hline
 & Train & Develop & Test & Total \\
\hline
\#Sentence  & 7,356 & 931 & 914 & 9,201 \\
\#Instance  & 12,247 & 1,624 & 1,614 & 15,485 \\
\#Entity  & 16,863 & 2,174 & 2,143 & 21,180 \\
\#Relation  & 12,247 & 1,624 & 1,614 & 15,485 \\
\#Image &  7,356 & 931 & 914 & 9,201 \\
\hline
\end{tabular}
\end{center}
\caption{ Summary of the dataset and splits in our experiments. `\#' denotes the number. 
}
\label{table:Statistics}
\vspace{-5mm}
\end{table}

\section{Experiment}

\vspace{-1mm}
\subsection{Setting}

We experiment with the MRE dataset\footnote{\url{https://github.com/thecharm/Mega}},
which contains 9,201 text-image pairs and 15,485 entity pairs with 23 relation categories.
The statistical information of the MRE dataset is listed in Table \ref{table:Statistics}.
Note that a sentence may contain several entity pairs, and thus a text-image pair can be divided into several instances, each with only one entity pair.
We follow the same split of training, development, and testing, as set in \citet{ZhengFFCL021}.
We compare our method with baselines in two categories: 
\textbf{1) Text-based RE methods} that traditionally leverage merely the texts of MRE data, including, \emph{BERT} \cite{devlin-etal-2019-bert}, \emph{PCNN} \cite{ZengLC015}, \emph{MTB} \cite{SoaresFLK19}, and \emph{DP-GCN} \cite{YuXZLWW20}.
\textbf{2) Multimodal RE methods} as in this work, including, \emph{BERT+SG} \cite{ZhengFFCL021}, \emph{MEGA} \cite{ZhengFFCL021}, 
\emph{VisualBERT} \cite{abs-1908-03557}, \emph{ViLBERT} \cite{LuBPL19},
\emph{RDS} \cite{0023HDWSSX22},
\emph{MKGformer} \cite{ChenZLDTXHSC22},
and \emph{HVPNet} \cite{ChenZLYDTHSC22}.

We use the pre-trained language-vision model CLIP (vit-base-patch32) to encode the visual and textual inputs.
We set the learning rate as 2e-5 for pre-trained parameters, and 2e-4 for the other parameters.
The threshold value $\lambda$ is set to 0.25; the temperature $\tau$ is 0.1; and $\beta$ is set to 0.01.
All the dimensions of node representations and GAT hidden sizes are set as 768-d.
We utilize the 2-order (i.e., $l=2$) context of each node to refine the nodes and edges of CMG.
For the latent topic modeling, we pre-define the number of topics as 10, and then we choose the Top-10 textual and visual keywords to enhance the semantic contexts of compressed CMG. 
All models are trained and evaluated using the NVIDIA A100 Tensor Core GPU.
Following existing MRE work, we adopt accuracy (Acc.), precision (Pre.), recall (Rec.), and F1 as the major evaluation metrics.

\begin{table}[!t]
\fontsize{9}{10.5}\selectfont
\setlength{\tabcolsep}{1.3mm}
\begin{center}
\resizebox{0.98\columnwidth}{!}{
\begin{tabular}{lcccc} 
\hline
  &   \bf Acc. & \bf Pre. & \bf   Rec. &  \bf F1  \\
\hline
\multicolumn{5}{l}{$\bullet$  \textbf{\emph{Text-based Methods}}} \\
BERT$^{\dag}$ & - & 63.85 & 55.79 & 59.55 \\
PCNN$^{\dag}$ & 72.67 & 62.85 & 49.69 & 55.49 \\
MTB$^{\dag}$  & 72.73 & 64.46 & 57.81 & 60.86 \\
DP-GCN$^{\flat}$   & 74.60 &  64.04 & 58.44 & 61.11\\
\hline
\multicolumn{5}{l}{$\bullet$  \textbf{\emph{Multimodal  Methods}}} \\
BERT(Text+Image)$^{\flat}$ & 74.59  & 63.07 & 59.53 & 61.25  \\
BERT+SG$^{\dag}$  & 74.09 & 62.95 & 62.65 & 62.80  \\
MEGA$^{\dag}$ & 76.15 & 64.51 & 68.44 & 66.41  \\
VisualBERT$_{base}^{\dag}$  & - & 57.15 & 59.48 & 58.30  \\
ViLBERT$_{base}^{\dag}$ & - & 64.50 & 61.86 & 63.16  \\
RDS$^{\dag}$  & - & 66.83   & 65.47 & 66.14  \\
HVPNeT$^{\dag}$ & - & \underline{83.64} & 80.78 & 81.85  \\
MKGformer$^{\dag}$  & \underline{92.31} & 82.67 & \underline{81.25} & \underline{81.95}  \\
\cdashline{1-5}
\bf Ours  & \textbf{94.06} &  \textbf{84.69} &  \textbf{83.38} &  \textbf{84.03} \\
\quad w/o \textsc{Gene} (Eq. \ref{GIB-total}) & 92.42 & 82.41 & 81.83 & 82.12 \\
\qquad w/o $I(\bm{z},G)$ (Eq. \ref{eq-GIB3}) &  93.64 & 83.61 & 82.34 & 82.97 \\
\quad w/o \textsc{Lamo} (Eq. \ref{eq-Lamo}) & 92.86 & 82.97 & 81.22 & 82.09 \\
\qquad w/o $\bm{o}^{T}$  & 93.05 &  83.95 & 82.53 & 83.23 \\
\qquad w/o $\bm{o}^{I}$ & 93.63 & 84.03 & 83.18 & 83.60 \\
\quad w/o VSG\&TSG  & {93.12} & 83.51 &82.67 & 83.09 \\
\quad w/o CMG & {93.97} & 84.38 & 83.20 & 83.78 \\
\hline
\end{tabular}
}
\vspace{-3mm}
\end{center}
\caption{Main results. 
`w/o $I(\bm{z},G)$' means \textsc{Gene} adjustment without GIB guidance.
`w/o CMG' means VSG and TSG are not connected with hyper-edge $E^{\times}$.
`w/o VSG\&TSG' means our method uses the embeddings of visual and text inputs without structural SG modeling.
Baselines with the superscript `$\dag$' are copied from their raw papers \cite{ChenZLDTXHSC22};
with `$\flat$' are from our re-implementation.
}
\label{Main results}
\vspace{-4mm}
\end{table}

\vspace{-1mm}
\subsection{Main Results}
\label{sec:Main Results}

Table \ref{Main results} shows the overall results.
First, compared to the traditional text-based RE, multimodal methods, by leveraging the additional visual features, exhibit higher performances consistently.
But without carefully navigating the visual information into the task, most MRE baselines merely obtain incremental improvements over text-based ones.
By designing delicate text-vision interactions, HVPNeT and MKGformer achieve the current state-of-the-art (SoTA) results.
Most importantly, our model boosts the SoTA with a very significant margin,
i.e., with improvements of 1.75\%(=94.06-92.31) in accuracy and 2.08\%(=84.03-81.95) in F1.
This validates the efficacy of our method.

\textbf{\begin{figure}[!t]
\centering
\includegraphics[width=1\columnwidth]{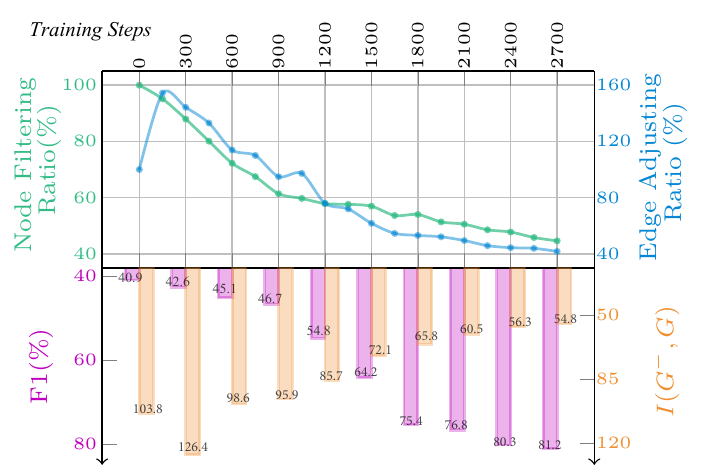}
\caption{
The trends of changing ratio of nodes and edges, along with the task performance and the mutual information between $G$ and $G^{-}$.
The model is without \textsc{Lamo}.
This is done on the developing set.
}
\label{fig:IBC1}
\vspace{-3mm}
\end{figure}}

\vspace{-3mm}
\paragraph{Model Ablation}
In the lower part of Table \ref{Main results}, we also study the efficacy of each part of our designs.
First of all, we see that both the \textsc{Gene} and \textsc{Lamo} modules show big impacts on the results, i.e., exhibiting a drop in F1 by 1.91\% and 1.94\% F1, respectively.
This confirms their fundamental contributions to the whole system.
More specifically, the GIB guidance is key to the information refinement in \textsc{Gene}, while both the textual and visual topic features are key to \textsc{Lamo}.
Also, it is critical to employ the SG for the structural modeling of the multimodal inputs.
And the proposal of the cross-modal graph is also helpful to task modeling.

\vspace{-2mm}
 \subsection{Analysis and Discussion}

\vspace{1pt}
To gain a deeper understanding of how our proposed methods succeed, we conduct further analyses to answer the following questions.

\vspace{3pt}
\noindent$\blacktriangleright$\textbf{RQ1}: \emph{Does \textsc{Gene} helps by really denoising the input features?}

\textbf{A}:
We first investigate the working mechanism of \textsc{Gene} on internal-information screening.
We plot the trajectories of the node filtering and the edge adjusting, during which we show the changing trends of overall performances and the mutual information $I(G^{-}, G)$ between the raw CMG ($G$) and the pruned one ($G^{-}$, i.e., $\bm{z}$).
As shown in Fig. \ref{fig:IBC1}, along with the training process both the number of nodes and edges decrease gradually, while the task performance climbs steadily as $I(G^{-}, G)$ declines.
These clearly show the efficacy of the task-specific information denoising by \textsc{Gene}.

\vspace{3pt}
\noindent$\blacktriangleright$\textbf{RQ2}: \emph{Are \textsc{Lamo} induced task-relevant topic features beneficial to the end task?}

\begin{table*}[!t]\em
\begin{center}
\resizebox{1.0\textwidth}{!}{
\begin{tabular}{cll}
\hline
  \textnormal{\textbf{Topic}} &  \textnormal{\textbf{Textual keywords}} & \textnormal{\textbf{Visual keywords (ID)}} \\
\hline
\#Politic & trump, president, world, new, china, leader, summit, meet, korean, senate  & \#1388, \#1068  \\
\#Music & tour, concert, video, live, billboard, album, styles, singer, taylor, dj  & \#1446, \#1891  \\
\#Love &  wife, wedding, engaged, ring, son, baby. girl, love, rose, annie &  \#434, \#1091  \\
\#Leisure & photo, best, beach, lake, island, bridge, view, florida, photograph, great  & \#679, \#895   \\
\#Idol & metgala, hailey, justin, taylor, rihanna, hit, show, annual, pope, shawn  & \#1021, \#352   \\
\#Scene & contain, near, comes, american, in,  spotted, travel, to, from, residents    &  \#535, \#167   \\
\#Sports & team, man, world, cup, nike, nba, football, join, play, chelsea & \#1700, \#109\\
\#Social & google, retweet, twitter, youtube, netflix, acebook, flight, butler, series, art  &  \#1043, \#1178\\
\#Show & show, presents, dress, interview, shot, speech, performing,  attend, portray, appear& \#477, \#930\\
\#Life & good, life, please, family, dog, female, people, boy, soon, daily & \#613, \#83\\
\hline\\
\end{tabular}
}
\includegraphics[width=1.0\textwidth]{img/vbow4.pdf}
\end{center}
\vspace{-3mm}
\caption{Top 10 key textual topic keywords and top 2 visual topic keywords discovered by \textsc{Lamo}.
}
\label{Topic keywords}
\end{table*}

\begin{figure}[!t]
\centering
\includegraphics[width=0.98\columnwidth]{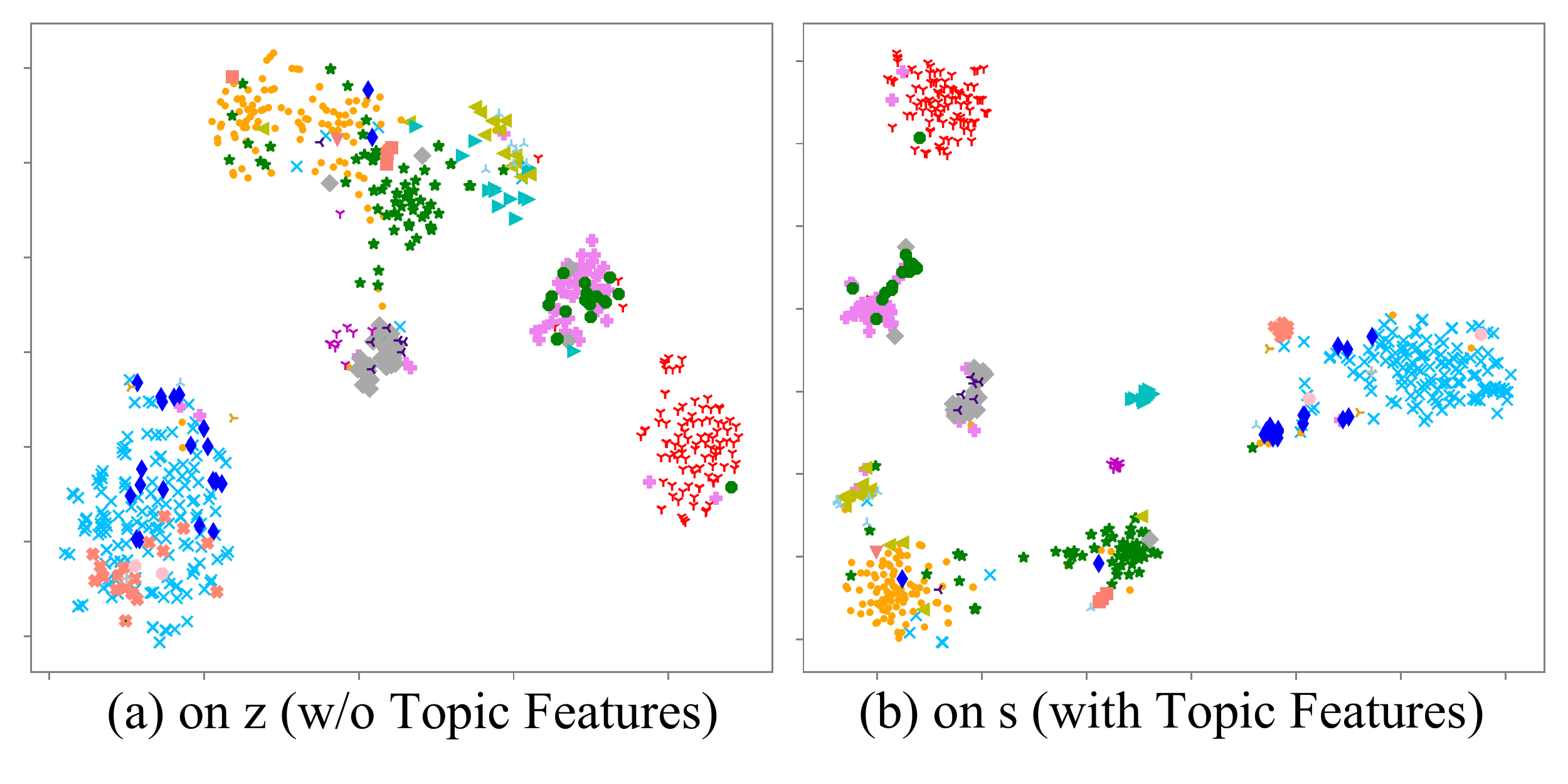}
\vspace{-1mm}
\caption{T-SNE visualization of the contextualized features with or without topic features.
Different colors indicate different ground-truth relation labels.
}
\label{fig:topicFortask}
\vspace{-3mm}
\end{figure}

\textbf{A}:
Now, we consider visualizing the learned contextual features in our system, including the $\bm{z}$ without integrating the topic features, and the $\bm{s}$ with rich topic information injected.
We separately project $\bm{z}$ and $\bm{s}$ into the ground-truth relational labels of the MRE task, as shown in Fig. \ref{fig:topicFortask}.
We see that both $\bm{z}$ and $\bm{s}$ have divided the feature space into several clusters clearly, thanks to the GIB-guided information screening.
However, there are still some wrongly-placed or entangled instances in $\bm{z}$, largely due to the input feature deficiency.
By supplementing more contexts with topic features, the patterns in $\bm{s}$ become much clearer, and the errors reduce.
This indicates that \textsc{Lamo} induces topic information beneficial to the task.

\begin{figure}[!t]
\centering
\includegraphics[width=0.8\columnwidth]{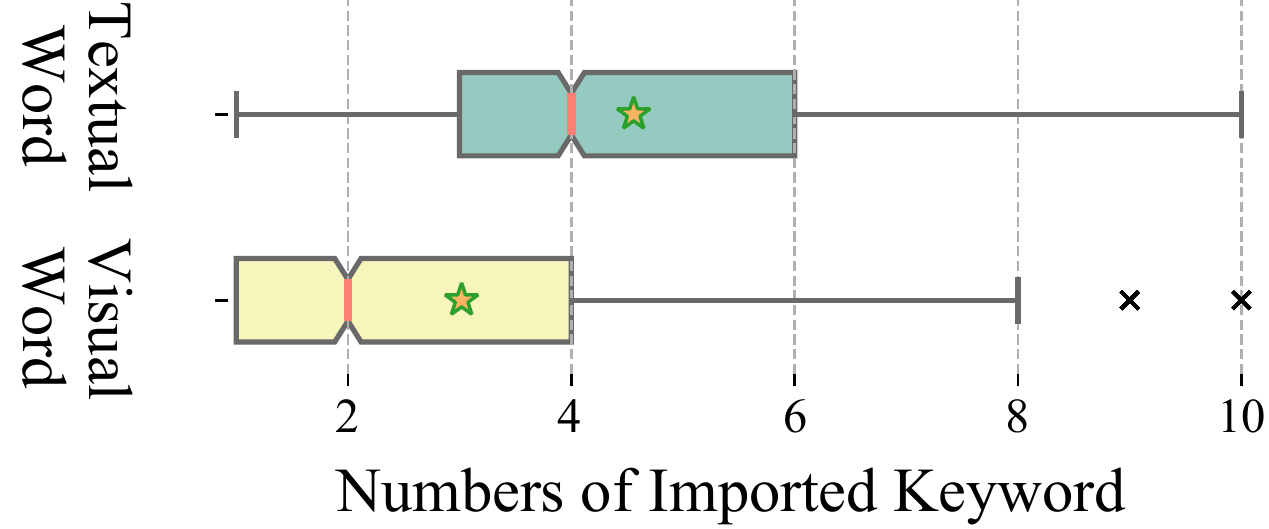}
\vspace{-1mm}
\caption{Distribution of numbers of textual and visual topic keywords imported for MRE.
}
\label{fig:ImportedTopicNumbers}
\vspace{-3mm}
\end{figure}

\begin{figure*}[!ht]
\centering
\includegraphics[width=0.98\textwidth]{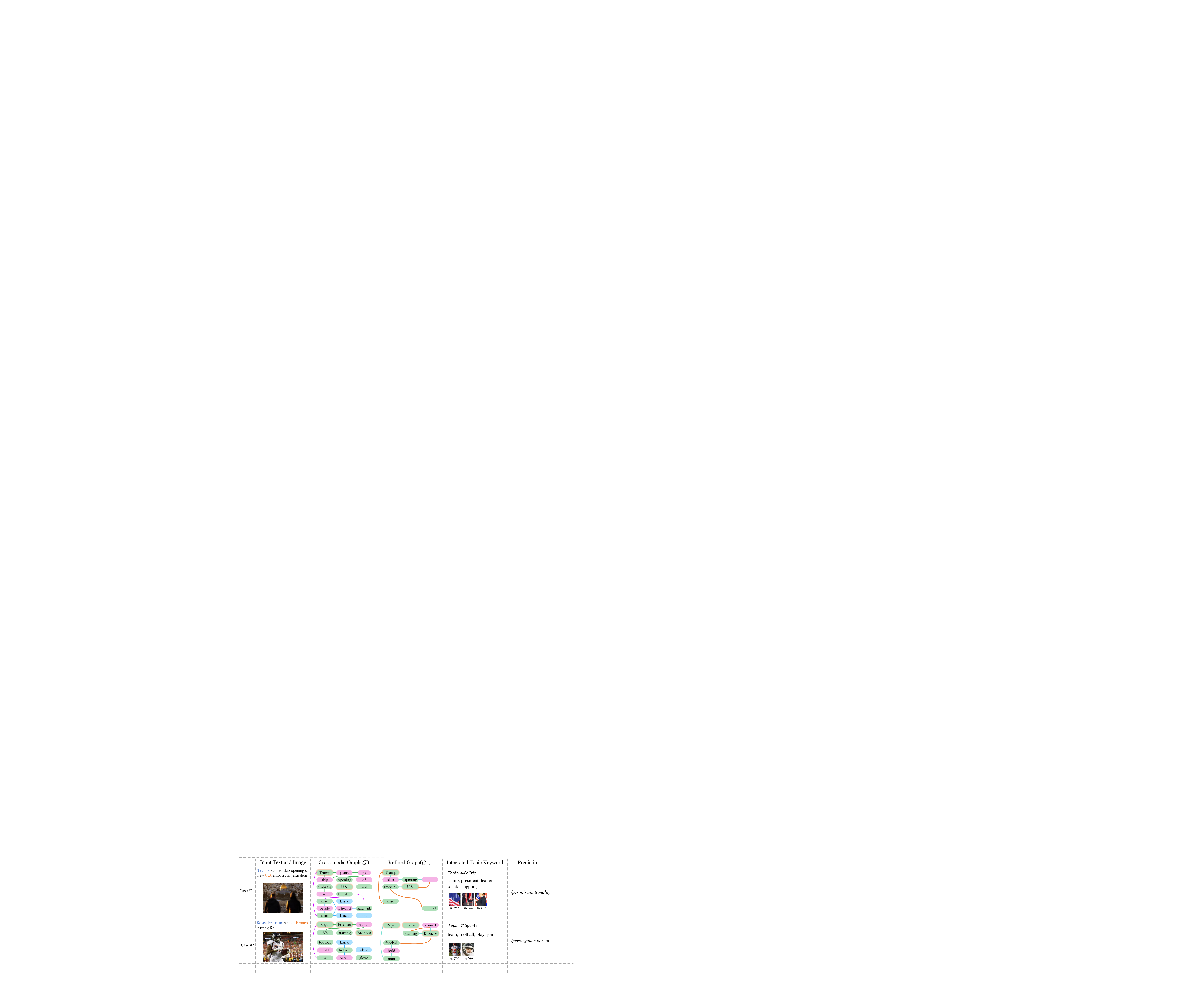}
\vspace{-1mm}
\caption{
Quantitative results of two testing examples, where our model made correct predictions.
}
\label{fig:case study}
\end{figure*}

\begin{figure}[!t]
\centering
\includegraphics[width=0.74\columnwidth]{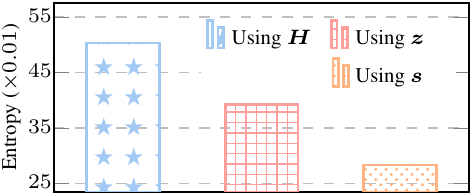}
\vspace{-1mm}
\caption{The entropy under various model settings.
}
\label{fig:colabration4}
\vspace{-5mm}
\end{figure}

Meanwhile, we demonstrate what latent topics \textsc{Lamo} can induce.
In Table \ref{Topic keywords} we show the top 10 latent topics with both the textual and visual keywords, where we notice that the latent topic information is precisely captured and modeled by \textsc{Lamo}.
Further, we study the variance of the latent topics in two modalities, exploring the different contributions of each type.
Technically, we analyze the numbers of the imported topic keywords of textual and visual ones respectively, by observing the attention weights $\alpha_{i}^{T/I}$ (Eq. \ref{att-weights}).
In Fig. \ref{fig:ImportedTopicNumbers} we plot the distributions.
It can be found that the model tends to make use of more textual contexts, compared with the visual ones.

\vspace{3pt}
\noindent$\blacktriangleright$\textbf{RQ3}: \emph{How do \textsc{Gene} and \textsc{Lamo} collaborate to solve the end task?}

\textbf{A}:
As demonstrated previously, the \textsc{Gene} is able to relieve the issue of noisy information, and \textsc{Lamo} can produce latent topics to offer additional clues for relation inference.
Now we study how these two modules cooperate together to reach the best results.
First, we use the learned feature $\bm{c}^{*}$ to calculate task entropy $-\sum p(Y|\bm{c}^{*})\log p(Y|\bm{c}^{*})$, where lower entropy means more confidence of the correct predictions.
We compute the entropy using $\bm{H}$ (initial context feature), using $\bm{z}$ (with denoised context feature) and using $\bm{s}$ (with feature denoising and topic enriched context), respectively, which represents the three stages of our system, as shown in Fig. \ref{fig:colabration4}.
As seen, after the information denoising and enriching by \textsc{Gene} and \textsc{Lamo} respectively, the task entropy drops step by step, indicating an effective learning process with the two modules.

We further empirically perform a case study to gain an intuitive understanding of how the two modules come to play.
In Fig. \ref{fig:case study} we illustrate the two testing instances, where we visualize the constructed cross-model graph structures, the refined graphs ($G^{-}$) and then the imported multimodal topic features.
We see that \textsc{Gene} has fine-grainedly removed those noisy and redundant nodes, and adjusted the node connections that are more knowledgeable for the relation prediction.
For example, in the refined graph, the task-noisy visual nodes, `\emph{man}' and textual nodes, `\emph{in}', `\emph{plans}' are removed, and the newly-generated edges (e.g., `\emph{Trump}'$\to$`\emph{US}', and `\emph{Broncos}'$\to$`\emph{football}') allow more efficient information propagation.
Also, the model correctly paid attention to the topic words retrieved from \textsc{Lamo} that are useful to infer the relation, such as `\emph{president}', `\emph{leader}' in case \#1, and \emph{team}', and `\emph{football}' in case \#2.

\begin{figure}[!t]
\centering
\includegraphics[width=0.96\columnwidth]{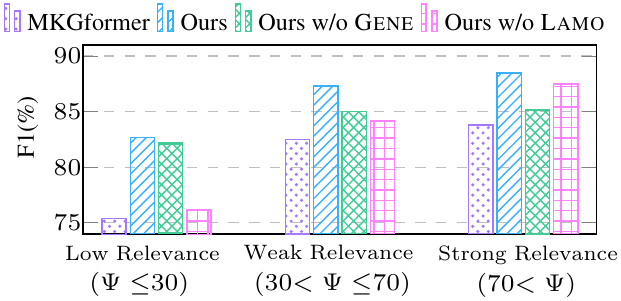}
\vspace{-1mm}
\caption{Results under varying text-image relevance.
}
\label{fig:relevance}
\vspace{-4mm}
\end{figure}

\vspace{3pt}
\noindent$\blacktriangleright$\textbf{RQ4}: \emph{Under what circumstances do the internal-information screening and external-information exploiting help?}

\textbf{A}:
In realistic scenarios, a wide range of multimodal tasks is likely to face the issues of internal-information over-utilization and external-information under-exploitation (or simultaneously).
Especially for the data collected from the online web, the vision and text pairs are not well correlated.
Finally, we take one step further, exploring when our idea of internal-information screening and external-information exploiting aids the tasks in such cases.
Technically, we first measure the vision-language relevance $\Psi$ of each image-text pair by matching the correspondence of the VSG and TSG structures.
And then, we group the instances by their relevance scores, and finally make predictions for different groups.
From Fig. \ref{fig:relevance}, it is observed that for the inputs with higher text-vision relevance, the \textsc{Gene} plays a greater role than \textsc{Lamo}, while under the case with less cross-modal feature relevance, \textsc{Lamo} contributes more significantly than \textsc{Gene}.
This is reasonable because most of the high cross-modal relevance input features come with rich yet even redundant information, where the internal-information screening is needed for denoising.
When the input text-vision sources are irrelevant, the exploitation of external features (i.e., latent topic) can be particularly useful to bridge the gaps between the two modalities.
On the contrary, \textbf{MKGformer} performs quite badly especially when facing with data in low vision-language relevance.
Integrating both the \textsc{Lamo} and \textsc{Gene}, our system can perform consistently well under any case.

\section{Related Works}

\vspace{-1.5mm}
As one of the key subtasks of the information extraction track, relation extraction (RE) has attracted much research attention \cite{YuXZLWW20,ChenZXDYTHSC22,TanHBN22,GuoKB23}.
The recent trend of RE has shifted from the traditional textual RE to the recent multimodal RE, where the latter additionally adds the image inputs in the former one for better performances, under the intuition that the visual information can offer complementary features to the purely textual input from other modalities.
\citet{ZhengWFF021} pioneers the MRE task with a benchmark dataset, which is collected from the social media posts that come with rich vision-language sources.
Later, more delicate and sophisticated methods are proposed to enhance the interactions between the input texts and images, and achieve promising results \citet{ZhengFFCL021,ChenZLYDTHSC22,ChenZLDTXHSC22}.

On the other hand, increasing attention has been paid to exploring the role of different information in the RE task.
As extensively revealed in prior RE studies, only a few parts of the input sentence can provide real clues for the relation inference \cite{xu-etal-2015-classifying,YuXZLWW20}, which inspires the proposal of textual feature pruning methods \cite{Zhang0M18,jin-etal-2022-supporting}.
More recently, \citet{vempala-preotiuc-pietro-2019-categorizing,VERE2022aaai} have shown that not always the visual inputs serve positive contributions in existing MRE models, as the social media data contains many noises.
\citet{0023HDWSSX22} thus introduce an instance-level filtering approach to directly drop out those images less-informative to the task.
However, such coarse-grained aggressive data deletion will inevitably abandon certain useful visual features.
In this work we propose screening the noisy information from both the visual and textual input features, in a fine-grained and more controllable manner, i.e., structure denoising via graph information bottleneck technique \cite{WuRLL20}.
Also, we adopt the scene graph structures to model both the vision and language features, which partially inherits the success from \citet{ZhengFFCL021} that uses visual scene graphs to represent input images.

Due to the sparse and noisy characteristics of social media data, as well as the cross-modal information detachment, MRE also suffers from feature deficiency problems.
We thus propose modeling the latent topic information as additional context features to enrich the inputs.
Multimodal topic modeling has received considerable explorations \cite{ChuZLWZH16,ChenWY0WL21}, which extends the triumph of the textual latent topic models as in NLP applications \cite{ZhuP0ZH20,FuBLJ20,XieHSA22}.
We however note that existing state-of-the-art latent multimodal models \cite{AnWLZ20,ZosaP22} fail to navigate the text and image into a unified feature space, which leads to irrelevant vision-text topic induction. 
We thus propose an effective latent multimodal model that learns coherent topics across two modalities.
To our knowledge, we are the first to attempt to integrate the multimodal topic features for MRE.

\section{Conclusion}
\vspace{-1.5mm}

In this paper, we solve the internal-information over-utilization issue and the external-information under-exploitation issue in multimodal relation extraction.
We first represent the input images and texts with the visual and textual scene graph structures, and fuse them into the cross-modal graphs.
We then perform structure refinement with the guidance of the graph information bottleneck principle.
Next, we induce latent multimodal topic features to enrich the feature contexts.
Our overall system achieves huge improvement over the existing best model on the benchmark data.
Further in-depth analyses offer a deep understanding of how our method advances the task.


\section*{Limitiation}

The main limitations of our work lie in the following two aspects: 
First, we take sufficient advantage of the scene graph (SG) structures, which are obtained by external SG parsers.
Therefore, the overall performance of our system is subject to the quality of the SG parser to some extent.
However, our system, by equipping with the refinement mechanism, is capable of resisting the quality degradation of SG parsers to a certain extent.
Second, the performance of the latent multimodal topic model largely relies on the availability of large-scale text-image pairs.
However, the size of the dataset of MRE is limited, which may limit the topic model in achieving the best effect.

\bibliography{custom}
\bibliographystyle{acl_natbib}

\newpage

\appendix

\section{Extended Method Specification}

\subsection{Scene Graph Generating}
\label{scene graph generating}

We mainly follow the prior practice of SG applications \cite{YangTZC19,GuJCZYW19} to acquire the visual scene graph (VSG) and textual scene graph (TSG).
A VSG or TSG contains three types of nodes, including the object, attribute, and relation nodes.

For VSG, we employ the FasterRCNN \cite{RenHGS15} as an object detector to obtain all the object nodes, 
and use MOTIFS \cite{ZellersYTC18} as a relation classifier to obtain the relation labels (nodes) as well as the relational edges, which is trained using the Visual Genome (VG) dataset \cite{krishna2017visual}.
We then use an attribute classifier to obtain attribute nodes.
For TSG generation, we first convert the sentences into dependency trees with a dependency parser, which is then transformed into the scene graph based on the rules defined at \citet{SchusterKCFM15}.
Note that the object nodes in VSG are image regions, while the object nodes in TSG are textual tokens.

\subsection{Node Embedding}
\label{visual node embedding}
In Section \ref{Scene Graph Generation}, we directly give the representations of nodes in VSG and TSG.
Here, we provide the encoding process in detail.

\paragraph{Visual Node Embedding}
In VSG, the visual feature vector of an object node is extracted from its corresponding image region; 
the feature of the attribute node is the same as its connected object, while the visual feature vector of a relationship node is extracted from the union image region of the two related object nodes.
Specifically, for each visual node, we first rescale it to 224-d $\times$ 224-d.
Subsequently, following \citet{DosovitskiyB0WZ21}, each visual node is split into a sequence of fixed-size non-overlapping patches $\{\bm{p}_k \in \mathbb{R}^{P \times P}\}$, where $P \times P$ is the patch size.
Then, we map all patches of $i$-th visual node to a $d$-dimensional vector $\bm{X}_i^{PC}$ with a trainable linear projection.
For each sequence of image patches, a [CLS] token embedding $\bm{x}_{CLS} \in \mathbb{R}^{d_1}$ is appended for the sequence of embedded patches, and an absolute position embeddings $X_i^{POS}$ also added to retain positional information.
The visual region of $i$-th node  is represented as:
\begin{equation}
\small
    \bm{Z}_i = [\bm{x}_{CLS};\bm{X}_i^{PC}] + X_i^{POS}\,,
\end{equation}
where $[;]$ denotes a concatenation.
Then, we feed the input matrix $\bm{Z}_i$ into the CLIP vision encoder to acquire the representation $\hat{\bm{x}}_i^{I}$.
Note that the [CLS] token is utilized to serve as a representation of an entire image region:
\begin{equation}
\small
    \hat{\bm{x}}_i^{I} = \text{CLIP}(\bm{Z}_i)_{[CLS]}\,.
\end{equation}
where $\hat{\bm{x}}_i^{I} \in \mathbb{R}^{d_1}$.
Since the category label of each node can provide the auxiliary semantic information, a label embedding layer is built to embed the word label of each node into a feature vector.
Given the one-hot vectors of the category label of each node, we first map it into an embedded feature vector $\bm{\bar{x}}_i^{I}$ by an embedding matrix $\bm{W}^{label} \in \mathbb{R}^{d_2 \times C_{label}}$, where is initialized by Glove embedding (i.e., $d_2=300$), $C_{label}$ is the number of categories.
And then, the embedding features of the category label corresponding to the node are fused to the visual features to obtain the final visual node embedding:
\begin{equation}
    \bm{x_i}^{I} = \text{Tanh}(\bm{W}_1[\hat{\bm{x}}_i^{I};\bm{\bar{x}}_i^{I}])\,.
\end{equation}
where $\bm{W}_1 \in \mathbb{R}^{d_1 \times (d_1+d_2)}$.

\paragraph{Textual Node Embedding}
In TSG, we utilize CLIP as the underlying encoder to yield the basic contextualized word representations for each textual node:
\begin{equation}
 \{\bm{x}_1^{T},\cdots,\bm{x}_m^{T}\}  = \text{CLIP}(\{v_1,\cdots,v_m\}) \,,
\end{equation}
where $\bm{x}_i^{T} \in \mathbb{R}^{d_1}$.

\subsection{Graph Encoding}
In Section \ref{Cross-modal Graph Construction} and Section \ref{GIB-guided Feature Refinement}, we introduce a graph attention model (GAT) to encode the cross-modal graph (CMG) and refined graph ($G^{-}$).
Here, we provide a detail. 
Technically, given a graph $G=(V, E)$, where $V$ is the set of nodes, and $E$ is the set of edges.
And the feature matrix $\bm{X} \in \mathbb{R}^{|V| \times d_1}$ of $V$ with $d_1$-dimensions.
The hidden state $\bm{h}_i$ of $i$-th  node will be updated as follows:
\begin{equation}\small
    \alpha_{i,j} = \frac{\exp(\text{LeakReLU}(\bm{W}_2 [\bm{x}_{i};\bm{x}_j]))}{\sum_{k \in \mathcal{N}(i)}  \exp(\text{LeakReLU}(\bm{W}_2 [\bm{x}_{i};\bm{x}_k]))} \,,
\end{equation}
\begin{equation}\small
         \bm{h}_i = \text{ReLU}(\sum\limits_{j}^{m+n} \alpha_{i,j} (\bm{W}_3  {\bm{x}}_j)) \,, 
\end{equation}
where $\mathcal{N}(i)$ denotes the neighbors of $i$-th node, $\bm{W}_2$ and $\bm{W}_3$ are learnable parameters.
In short, we denote the graph encoding as follows:
\begin{equation}
    \small
    \bm{H} = \text{GAT}(G, \bm{X})\,.
\end{equation}

\begin{figure}[!ht]
\centering
\includegraphics[width=0.85\columnwidth]{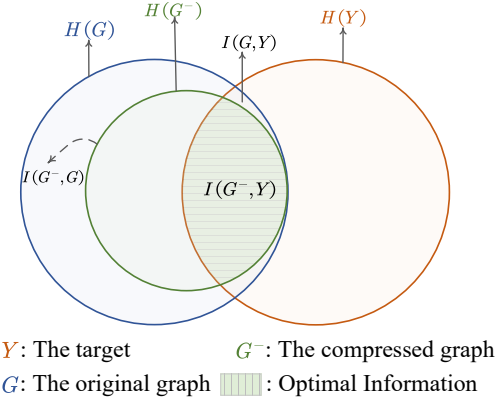}
\caption{The Venn diagram visualization of GIB.
}
\label{fig:gib-venn}
\end{figure}

\subsection{Detailed GIB-guided Feature Refinement}
\label{GIB-process}

\paragraph{Introduction to GIB}
\label{Introduction of GIB}
Here, we provide more background information about the GIB principle.
Given the original graph $G$, and the target $Y$, the goal of representation learning is to obtain the compressed graph ${G}^{-}$ which is maximally informative w.r.t $Y$ (i.e., sufficiency, $I({G}, Y) = I({G}^{-}, Y)$), and without any noisy information (i.e., minimality, $I({G}^{-}, {G}) - I({G}^{-}, Y) = 0$), as indicated in Fig. \ref{fig:gib-venn}.
To encourage the information compressing process to focus on the target information, GIB was proposed to enforce an upper bound $I_c$ to the information flow from the original graph to the compressed graph, by maximizing the following objectives:
\begin{equation}\label{GIB7}
\setlength\abovedisplayskip{2pt}
\setlength\belowdisplayskip{2pt}
\small
    \mathop{max}\limits_{{G}^{-}} I({G}^{-}, Y)\; s.t.\:I({G}^{-}, {G}) \leq I_c \,.
\end{equation}
Eq. (\ref{GIB7}) implies that a compressed graph can improve the generalization ability by ignoring irrelevant distractors in the original graph. 
By using a Lagrangian objective, GIB allows the ${G}^{-}$ to be maximally expressive about $Y$ while being maximally compressive about ${G}$ by:
\setlength\abovedisplayskip{2pt}
\setlength\belowdisplayskip{2pt}
\begin{equation}\small
\label{GIB8}
    \mathop{max}\limits_{{G}^{-}} I({G}^{-}, Y) - \beta I({G}^{-}, {G})\,,
\end{equation}
where $\beta$ is the Lagrange multiplier.
For the sake of consistency with the main body of the paper, the objective can be rewritten to:
\setlength\abovedisplayskip{2pt}
\setlength\belowdisplayskip{2pt}
\begin{equation}\small
\label{GIB9}
    \mathop{min}\limits_{{G}^{-}} -I({G}^{-}, Y) + \beta I({G}^{-}, {G})\,.
\end{equation}

However, the GIB objective in Eq. (\ref{GIB9}) is notoriously hard to optimize due to the intractability of mutual information and the discrete nature of irregular graph data.
By assuming that there is no information loss in the encoding process \cite{Tian0PKSI20}, the graph representation $\bm{z}$ of ${G}^{-}$ is utilized to optimize the GIB objective in Eq. (\ref{GIB}), leading to $ -I({G}^{-}, Y) \sim -I(\bm{z}, {G})$, $ I({G}^{-}, {G}) \sim I(\bm{z}, Y)$.
Therefore, the Eq. (\ref{GIB9}) can be computed as:
\setlength\abovedisplayskip{2pt}
\setlength\belowdisplayskip{2pt}
\begin{equation}\small
\label{R-GIB}
    -I({G}^{-}, Y) + \beta I({G}^{-}, {G}) \sim -I(\bm{z}, Y) + \beta I(\bm{z}, {G}) \,.
\end{equation}

\begin{figure}[!ht]
\centering
\includegraphics[width=0.95\columnwidth]{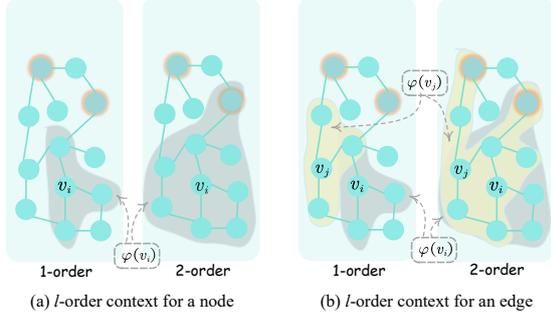}
\caption{The $l$-order context for a node and an edge.
}
\label{fig:l_order}
\end{figure}

\paragraph{Attention Operation for Node Filtering and Edge Adjusting}
In Section \ref{GIB-guided Feature Refinement}, we utilize the $l$-order context to determine whether a node should be filtered or an edge should be adjusted since the nodes and edges in a graph have local dependence, as shown in Fig. \ref{fig:l_order}.
Here, we give a detail of the calculation for the Att($\cdot$) operation in Eq. (\ref{pi-v}) and Eq. (\ref{pi-e}).
In Eq. (\ref{pi-v}), the attention operation can be computed as:
\begin{equation}\small
    \begin{aligned}
        \alpha_{i,k}^{v} &= \frac{\exp(\bm{W}_4 [\bm{h}_i; \bm{h}_k])}{\sum_{c \in \Phi(\{v_i, \varphi(v_i)\})} \exp(\bm{W}_4 [\bm{h}_i; \bm{h}_c])} \\
        \bm{r}_{i}^{v} &= \text{Tanh}(\sum\limits_{k \in \Phi(\{v_i, \varphi(v_i)\})} \alpha_{i, k}^{v} (\bm{W}_5 \bm{h}_k))
    \end{aligned}
\end{equation}
where $\Phi(\{v_i, \varphi(v_i)\})$ is a function to retrieve the index of a node in a set.
Similarly, we consider the $l$-order context to calculate the $\bm{r}^{e}_{i, j}$ in Eq.(\ref{pi-e}):
\begin{equation}
    \small
    \begin{aligned}
        \alpha_{i,j,k}^{e} &= \frac{\exp(\bm{W}_6 [\bm{h}_i; \bm{h}_j;\bm{h}_k])}{\sum\limits_{c \in \Phi(\{v_i, \varphi(v_i), v_j, \varphi(v_j)\})} \exp(\bm{W}_6 [\bm{h}_i; \bm{h}_j; \bm{h}_c])} \\
        \bm{r}_{i,j}^{v} &= \text{Tanh}(\sum\limits_{k \in \Phi(\{v_i, \varphi(v_i), v_j, \varphi(v_j)\})} \alpha_{i, j, k}^{e} (\bm{W}_7\bm{h}_k))
    \end{aligned}
\end{equation}

\paragraph{Detailed GIB Optimization}
\label{Detailed GIB Optimization}
First, we examine the second term $I(\bm{z}, {G})$ in Eq. (\ref{GIB-total}).
Same as \citet{Sun0P0FJY22}, we employ variational inference to compute a variational upper bound for $I(\bm{z}, {G})$
as follow:
\begin{equation}\label{GIB3}
\small
  I(\bm{z}, {G}) \leq \int p(\bm{z}|{G})\log\frac{p(\bm{z}|{G})}{r(\bm{z})}d\bm{z}d{G} \,,
\end{equation}
where $r(\bm{z})$ is the variational approximation to the prior distribution $p(\bm{z})$ of $\bm{z}$, which is treated as a fixed $d_1$-dimensional spherical Gaussian as in \citet{AlemiFD017}, i.e., $r(\bm{z}) = {N}(\bm{z}|0, \bm{I})$.
We use reparameterization trick (\cite{KingmaW13}) to sample $\bm{z}$ from the latent distribution according to $p(\bm{z}|{G})$, i.e., $p(\bm{z}|{G}) = {N}(\bm{\mu}_z, \bm{\sigma}_z)$,
where $\bm{\mu}_z$ and $\bm{\sigma}_z$ is the mean vector and the diagonal co-variance matrix of $\bm{z}$, which can be computed as:
\begin{equation}\small
    \bm{\mu}_{z} = \text{FFN}(\bm{a}) \,; \bm{\sigma}_z = \text{Softplus}(\text{FFN}(\bm{a}) \,,
\end{equation}
where $\bm{a}$ is the context feature of $G^{-}$ obtained from Eq.(\ref{final-context-feature-a}).
$\bm{z}$ is sampled by $\bm{z}= \bm{\mu}_{z} + \bm{\sigma}_z \cdot \varepsilon $, where $\varepsilon \in {N}(0, \bm{I})$.
We could reach the following optimization to approximate $I(\bm{z}, {G})$:
\begin{equation}\label{GIB4}\small
     I(\bm{z}, {G}) = \text{KL}(p(\bm{z}|{G})||r(\bm{z})) \,,
\end{equation}
where $KL(\cdot || \cdot)$ is the Kullback Leibler (KL) divergence \cite{HersheyO07}.

Then, we examine the first term in Eq. (\ref{GIB-total}), which encourages $\bm{z}$ to be informative to $Y$.
We expand $I(\bm{z}, Y)$ as:

\begin{small}
\begin{equation}\label{GIB10}
\begin{aligned}
    -I(\bm{z}, Y) &\leq -\! \int\! p(Y, \bm{z})\log q(Y|\bm{z}) dY d\bm{z} + H(Y) \\
    & := \mathcal{L}_{\text{\scriptsize CE}}(q(Y|\bm{z}), Y) \,,
\end{aligned}
\end{equation}
\end{small}
where $q(Y|\bm{z})$ is the variational approximation of the true posterior $p(Y, \bm{z})$.
Eq. (\ref{GIB10}) indicates that minimizing $-I(\bm{z}, Y)$ is achieved by minimization of the classification loss between $Y$ and $\bm{z}$, we model it as an MLP classifier with parameters.
The MLP classifier takes $\bm{z}$ as input and outputs the predicted label.

\begin{figure}[!ht]
\centering
\includegraphics[width=0.95\columnwidth]{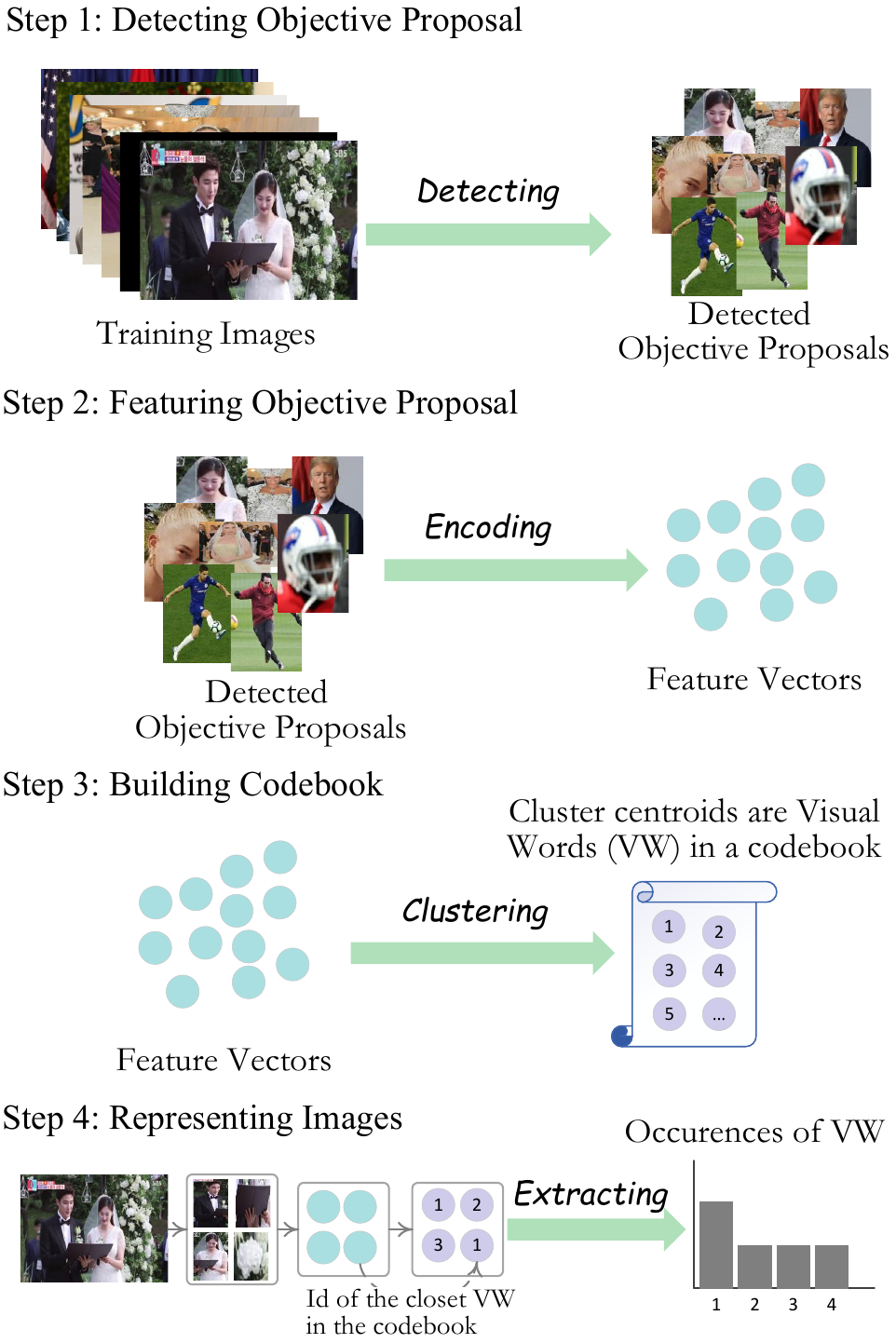}
\caption{The extraction process of visual BoW features.
}
\label{fig:vw-build}
\end{figure}

\subsection{Detailed Latent Multimodal Topic Modeling}
\label{Latent Multimodal Topic Modeling}

\paragraph{Visual BoW Feature Extraction}
\label{Visual Word Generating}
As mentioned in Section \ref{sec:Latent Multimodal Topic Modeling}, we represent image $I$ with visual BoW (VBoW) features.
Here, we introduce how to extract VBoW features from an image.
We compute the objective-level visual words in the following four steps, as shown in Fig. \ref{fig:vw-build}:
\begin{itemize}
    \item \textbf{Step 1: Detecting Objective Proposal}. We first employ a Faster-RCNN \cite{RenHGS15} as an objective detector to extract all the objective proposals in the training dataset.
    \item \textbf{Step 2: Featuring Objective Proposal}: We use a pre-trained vision language model to obtain the feature descriptors (vectors) of each objective proposal.
    \item \textbf{Step 3: Building the Codebook}: After obtaining the feature vectors, these feature vectors are clustered by a kmeans algorithm, where the number of clusters is set to 2,000. Cluster centroids are taken as visual words.
    \item \textbf{Step 4: Representing Images}: Similar to the extraction of Bag-of-word (BoW) features for text representation, we build the Visual Bag-of-Word (VBoW) features for images. Specifically, using this codebook, each feature vector of the objective proposal in an image is replaced with the id of the nearest learned visual word.
\end{itemize}

\paragraph{Detailed Latent Topic Modelling Optimization}
In Section \ref{sec:Latent Multimodal Topic Modeling}, we directly provide the optimal objective.
In the following, we introduce how to optimize $\textsc{Lamo}$ concretely. 
First of all, the prior parameters of $\bm{\theta}, \bm{\mu}$ and $\bm{\sigma}$ are estimated from the input data and defined as:
\begin{equation}
\small
    \bm{\mu} = f_\mu(f(\bm{H})),\; \text{log}\bm{\sigma} = f_\sigma(f(\bm{H}))\,,
\end{equation}
where $\bm{H}$ is the contextualized representation obtained from CMG, $f(\cdot)$ is an aggregation function, and $f_*(\cdot)$ is a neural perceptron that linearly transforms inputs, activated by a non-linear transformation.
Note that we can generate the latent topic variable $\bm{\varpi}$ from $p(\bm{\theta}|T, I)$ by sampling, i.e., $\bm{\varpi}=\bm{\mu} + \bm{\sigma} \cdot \bm{\varepsilon}$, where $\bm{\varepsilon} \in \mathcal{N}(0, \bm{I})$.
Then we employ Gaussian softmax to draw topic distribution $\bm{\theta}$:
\begin{equation}
    \small
    \bm{\theta }= \text{Softmax}(\text{FFN}(\bm{\varpi}))    
\end{equation}

Similar to previous neural topic models only for handling text \cite{BianchiTHNF21}, we consider autoregressively reconstructing the textual and visual BoW features of input by learned topic distribution $\bm{\theta}$:
\setlength\abovedisplayskip{2pt}
\setlength\belowdisplayskip{2pt}
\begin{equation}\small
p(\bm{b}_i^T|\bm{\chi}, \bm{\theta}) = \text{Softmax}(\bm{\theta} \cdot \bm{\chi} | \bm{b}_{<i}^T )\,,
\end{equation}
\begin{equation}\small
p(\bm{b}_i^I|\bm{\psi}, \bm{\theta}) = \text{Softmax}(\bm{\theta} \cdot \bm{\psi} | \bm{b}_{<i}^I)\,.
\end{equation}
The objective function of latent multimodal topic modeling is to maximize the evidence lower bound (ELBO), as derived as follows:
\setlength\abovedisplayskip{2pt}
\setlength\belowdisplayskip{2pt}
\begin{equation}\small
\begin{aligned}
    \mathcal{L}_{LAMO} = &KL(q(\bm{\theta})||p(\bm{\theta}|T, I)) \\
    &- \mathbb{E}_{q(\bm{\theta})}[p(\bm{b}^{T}|\bm{\theta}, \bm{\chi})] \\
    &- \mathbb{E}_{q(\bm{\theta})}[p(\bm{b}^{I}|\bm{\theta}, \bm{\psi})] \\
    =&\mathcal{L}_{KL} + \mathcal{L}_{RecT} + \mathcal{L}_{RecI}\,,
\end{aligned}
\end{equation}
where $q(\bm{\theta})$ is the prior probability of $\bm{\theta}$, set as a standard Normal prior $\mathcal{N}(0, \bm{I})$.

\section{Extended Experiments Setting}
\label{Extend Experiments Setting}

\subsection{Baselines}
We compare our model with two categories of baseline systems.
\paragraph{Text-based Methods,} which only leverage the texts of MRE data. 
\begin{itemize}
        \item \textbf{BERT} \cite{devlin-etal-2019-bert} is only fine-tuned on the dateset by \citet{ZhengFFCL021}.
        \item \textbf{PCNN} \cite{ZengLC015} leverages external knowledge graphs to extract relations in a distantly supervised manner, which is employed in MRE dataset by \citet{ZhengFFCL021}.
        \item \textbf{MTB} \cite{SoaresFLK19} is a RE-oriented pretraining model based on BERT, which is applied in MRE dataset by \citet{ZhengFFCL021}.
        \item \textbf{DP-GCN} \cite{YuXZLWW20} propose dynamical pruning GCN for relation extraction, we re-implemented the framework and apply it to the MRE dataset.
    \end{itemize}

\paragraph{Multimodal Methods,} which utilize the additional visual information to enhance the textual RE.
\begin{itemize}
        \item \textbf{BERT+SG} \cite{ZhengFFCL021} simply concatenate the textual representation with visual features extracted. 
        \item \textbf{MEGA} \cite{ZhengFFCL021} leverage the alignment between textual and visual graphs to learn better semantic representation for MRE.
        \item \textbf{VisualBERT} \cite{abs-1908-03557} is a single-stream structure via self-attention to discover implicit alignments between language and vision, which is then fine-tuned on the MRE dataset by \citet{ChenZLDTXHSC22}. 
        \item \textbf{ViLBERT} \cite{LuBPL19} consider employing two parallel streams for visual and language processing, which is then fine-tuned on the MRE dataset by \citet{ChenZLDTXHSC22}.
        \item \textbf{HVPNet} \cite{ChenZLYDTHSC22} propose to incorporate visual features into each self-attention layer of BERT.
        \item \textbf{MKGformer} \cite{ChenZLDTXHSC22} introduce a hybrid transformer architecture, in which the underlying two encoders are utilized to capture basic textual and visual features, and the upper encoder to model the interaction features between image and text.
        \item \textbf{RDS} \cite{0023HDWSSX22} design a data discriminator via reinforcement learning to determine whether data should utilize additional visual information for the relation inference.
    \end{itemize}

\subsection{Calculating Text-image Relevance}

In Fig. \ref{fig:relevance} we measure the relevance of input text-image pairs.
Technically, we adopt the CLIP model to yield a vision-language matching score.
Instead of directly feeding the whole picture and sentence into CLIP, we take a finer-grained method.
Because in the MRE data, the picture and sentence pair collected from social media sources comes with low correlations, and if directly measuring their relevance at the instance level, our preliminary experiment shows that the highest text-image relevance score by CLIP is only 45\%.
Thus, we measure the picture and sentence pair by matching their correspondence of the VSG and TSG structures. 
We take their object nodes and the attribute nodes at the treatment targets, and calculate the vision-language pairs with CLIP at the node level:
\begin{equation}\small
 \Psi(I,T) =  \frac{1}{Z} \sum_{i,j}  \text{CLIP} (\bm{x}^I_i,\bm{x}^T_j |  G^I, G^T) \,,
\end{equation}
where $Z$ is the normalization term.

\end{document}